%% file: mypaper.tex
\documentclass[sigconf]{acmart}

\usepackage{array}
\usepackage{multirow}    
\usepackage{makecell}    
\usepackage{xcolor}      
\usepackage{booktabs}    
\usepackage[table]{xcolor}
\usepackage{arydshln}    
\usepackage{placeins}    
\usepackage{enumitem}

\AtBeginDocument{%
  }

\setcopyright{acmlicensed}
\copyrightyear{2018}
\acmYear{2018}
\acmDOI{XXXXXXX.XXXXXXX}
\acmConference[Conference acronym 'XX]{Make sure to enter the correct
  conference title from your rights confirmation email}{June 03--05,
  2018}{Woodstock, NY}
\acmISBN{978-1-4503-XXXX-X/2018/06}

\begin{document}

\title{SHRP: Specialized Head Routing and Pruning for Efficient Encoder Compression}

\author{Zeli Su\textsuperscript{*}}
\email{rickamorty@muc.edu.cn}
\affiliation{%
  \institution{Minzu University of China}
  \city{Beijing}
  \country{China}
}

\author{Ziyin Zhang\textsuperscript{*}}
\email{daenerystargaryen@sjtu.edu.cn}
\affiliation{%
  \institution{Shanghai Jiao Tong University}
  \city{Shanghai}
  \country{China}
}

\author{Wenzheng Zhang\textsuperscript{*}}
\email{zwz@stu.pku.edu.cn}
\affiliation{%
  \institution{Peking University}
  \city{Beijing}
  \country{China}
}

\author{Zhou Liu}
\email{2501111514@stu.pku.edu.cn}
\affiliation{%
  \institution{Peking University}
  \city{Beijing}
  \country{China}
}

\author{Guixian Xu\textsuperscript{†}}
\email{guixian_xu@muc.edu.cn}
\affiliation{%
  \institution{Minzu University of China}
  \city{Beijing}
  \country{China}
}

\author{Wentao Zhang\textsuperscript{†}}
\email{wentao.zhang@pku.edu.cn}
\affiliation{%
  \institution{Peking University}
  \city{Beijing}
  \country{China}
}

\renewcommand{\shortauthors}{Su et al.}
\thanks{* Equal contribution.}
\thanks{† Corresponding authors.}

\renewcommand{\shortauthors}{Trovato et al.}

\begin{abstract}
Transformer encoders are widely deployed in large-scale web services for natural language understanding tasks such as text classification, semantic retrieval, and content ranking. However, their high inference latency and memory consumption pose significant challenges for real-time serving and scalability. These limitations stem largely from architectural redundancy, particularly in the attention module. The inherent parameter redundancy of the attention mechanism, coupled with the fact that its attention heads operate with a degree of independence, makes it particularly amenable to structured model compression. In this paper, we propose \textbf{SHRP (Specialized Head Routing and Pruning)}, a novel structured pruning framework that automatically identifies and removes redundant attention heads while preserving most of the model's accuracy and compatibility. SHRP introduces \textit{Expert Attention}, a modular design that treats each attention head as an independent expert, followed by a lightweight shared expander feed-forward network that refines their outputs. The framework employs a unified Top-1 usage-driven mechanism to jointly perform dynamic routing during training and deterministic pruning at deployment. Experimental results on the GLUE benchmark using a BERT-base encoder show that SHRP achieves \textbf{93\%} of the original model accuracy while reducing parameters by \textbf{48\%}. Under an extreme compression scenario where \textbf{11/12 of the layers are pruned}, the model still maintains \textbf{84\%} accuracy and delivers a \textbf{4.2}$\times$ throughput gain while reducing computation to as low as \textbf{11.5\%} of the original FLOPs, demonstrating its practical utility for large-scale and latency-sensitive web deployments.
\end{abstract}





\maketitle

\input{content/sec-1-introduction}

\input{content/sec-2-Related-work}

\input{content/sec-3-Methodology}

\input{content/sec-4-Experiment}

\section{Conclusion}
In this work, we presented \textbf{SHRP} (\textit{Specialized Head Routing and Pruning}), a structured pruning framework for efficient Transformer encoders. At its core, SHRP replaces standard multi-head attention with \textbf{Expert Attention}, where each head is modularized as a specialized expert paired with a shared lightweight feed-forward branch. This design enables experts to specialize during training and later be pruned cleanly based on usage statistics. 

We combined this building block with a progressive two-stage training strategy and a unified Top-1 routing–pruning mechanism, producing compact and deterministic models. Extensive evaluation on the GLUE benchmark shows that SHRP (i) \textbf{outperforms traditional head pruning methods} in both compression ratio and runtime efficiency, (ii) \textbf{achieves smooth accuracy–efficiency trade-offs} under aggressive pruning, and (iii) \textbf{provides interpretable pruning behavior}, with clear expert specialization emerging from Top-1 routing. 

By aligning architectural modularization with pruning and deployment needs, SHRP demonstrates that pruning can serve not only as a model optimization tool but also as a framework for practical systems design. The result is fast, compact, and accurate Transformer encoders that are well-suited for large-scale classification and probing tasks.

\clearpage
\bibliographystyle{ACM-Reference-Format}
\bibliography{citation}

\input{content/appendix}

\end{document}

%% file: content/sec-1-introduction.tex
\section{Introduction}

Transformer-based architectures—spanning both encoder-only and decoder-only designs—form the backbone of modern NLP. Encoder models such as BERT~\cite{Bert} and XLM-R~\cite{xlm-r} are widely used in classification, retrieval, and tagging pipelines~\cite{wang2019glue,Wang2019superglue}, while decoder-only models such as LLaMA~\cite{llama}, DeepSeek~\cite{deepseek}, and Qwen~\cite{qwen2} have driven rapid progress in generative applications. With their ability to capture long-range dependencies and scale to billions of parameters, Transformers have become indispensable across a wide range of tasks.

However, despite their representational power, encoder deployments remain inefficient at production scale. Inference over millions of user queries and documents—common in large-scale web services for natural language understanding tasks such as text classification, semantic retrieval, and content ranking—pushes latency, memory, and energy consumption to their limits. Even on modern GPUs, encoder models are substantially overparameterized: prior work has shown that many attention heads contribute little to final predictions~\cite{michel2019sixteen,voita2019analyzing}, and that a large fraction of model parameters and FLOPs can be removed without degrading performance~\cite{hou2020dynabert,sanh2020movement}. This indicates a fundamental gap between model capacity and actual usage—a form of architectural redundancy that motivates compression.

Redundancy arises at multiple levels: first, \textbf{at the head level}, where many attention heads are inactive or underutilized; second, \textbf{at the feed-forward level}, where the FFN sublayer accounts for the majority of both parameter count and computation cost. Yet most existing methods fail to fully exploit this redundancy. \textbf{Coarse-grained approaches}, such as layer dropping or head pruning~\cite{michel2019sixteen,voita2019analyzing}, remove large components, but tend to suffer from sharp accuracy loss under aggressive pruning. \textbf{Fine-grained methods}, including movement pruning and structured sparsity~\cite{sanh2020movement,zafrir2021prunebert}, achieve higher compression but often rely on heuristic scoring, custom masking, or hardware-aware patterns that limit real-world deployment. Critically, most attention-head pruning methods operate only within the attention sublayer. The FFNs—despite dominating overall computation—are typically left untouched. To preserve dimensionality, many approaches apply zero-padding or weight replication, which preserves FLOPs and fails to reduce runtime.

These limitations motivate us to rethink structured compression from first principles. \textbf{This paper proposes SHRP (Specialized Head Routing and Pruning)}, a new pruning framework that compresses encoder models by modularizing attention heads as independent experts and coupling them with a shared lightweight Expander-FFN. Rather than relying on handcrafted heuristics, SHRP draws inspiration from Mixture-of-Experts (MoE) models~\cite{shazeer2017outrageously,lepikhin2021gshard}, but repurposes the mechanism for compression instead of capacity expansion. By introducing lightweight routing during training, SHRP enables interpretable head usage patterns that guide pruning—without introducing any dynamic behavior at inference.

The overview of the SHRP framework shown in \textbf{Figure~\ref{fig:shrp-pipeline}} . The process begins with \emph{Attention Transformation}, where standard multi-head attention is replaced by \textbf{Expert Attention}—a modular structure where each head is routed independently through a shared lightweight \textbf{Expander FFN}. The framework then proceeds through four phases: layer-wise conversion with balanced routing, expert specialization, head-usage analysis, and final Top-1 pruning.

\textbf{Key challenges.} Recasting heads as experts raises several new challenges:  
(i) how to maintain expert diversity during early training to avoid collapse;  
(ii) how to induce interpretable and stable head usage for pruning;  
(iii) how to integrate FFN compression without disrupting residual connections; and  
(iv) how to eliminate the routing overhead, which can account for up to 50\% of runtime once the model is slimmed down.

\textbf{Our solution.} SHRP introduces three innovations. First, we replace standard multi-head attention with \textbf{Expert Attention}, where each head is modularized and routed independently through a shared \textbf{Expander FFN} that performs a direct $d_{\text{head}} \to d$ projection. This allows both attention and FFN sublayers to be pruned jointly. Second, we adopt a \textbf{progressive two-stage training pipeline}: a load-balancing objective encourages uniform usage in early layers, followed by a specialization phase that yields clear pruning signals. Third, we employ \textbf{Top-1 routing and pruning}, where usage statistics are tracked during training and underused experts are pruned post hoc. The router is then removed, yielding a static encoder with no dynamic computation. While the SHRP framework can in principle generalize to Transformer variants, we focus on encoder-only architectures due to their practical dominance in deployment and their suitability for structured pruning (detail discussing in Appendix~\ref{appendix:why-encoders}).


\textbf{Contributions.} In summary, this paper makes the following contributions:
\begin{itemize}
    \item We introduce \textbf{SHRP} (\textit{Specialized Head Routing and Pruning}), a unified pruning framework that compresses encoder models by treating attention heads as independent experts routed through a shared Expander-FFN, enabling joint pruning of attention and FFN components.
    \item We propose a \textbf{two-stage progressive training pipeline} that induces interpretable head specialization, allowing pruning without auxiliary scoring functions or architectural constraints.
    \item We demonstrate that \textbf{Top-1 routing and pruning} achieves the best trade-off between accuracy and compression, and removes all routing overhead at inference time, producing compact and deterministic encoders.
\end{itemize}

On the GLUE benchmark, SHRP achieves up to \textbf{88.5\% parameter reduction}, \textbf{4.2$\times$ throughput improvement}, and reduces compute to as low as \textbf{11.5\% of original FLOPs}, while retaining over \textbf{84\%} of baseline accuracy even under aggressive compression. These results demonstrate that modular design, combined with MoE-inspired training and principled routing, enables pruning to move beyond heuristics toward deployment-ready encoder optimization.

%% file: content/sec-2-Related-work.tex
\begin{figure*}[t]
    \centering
    \includegraphics[width=\textwidth]{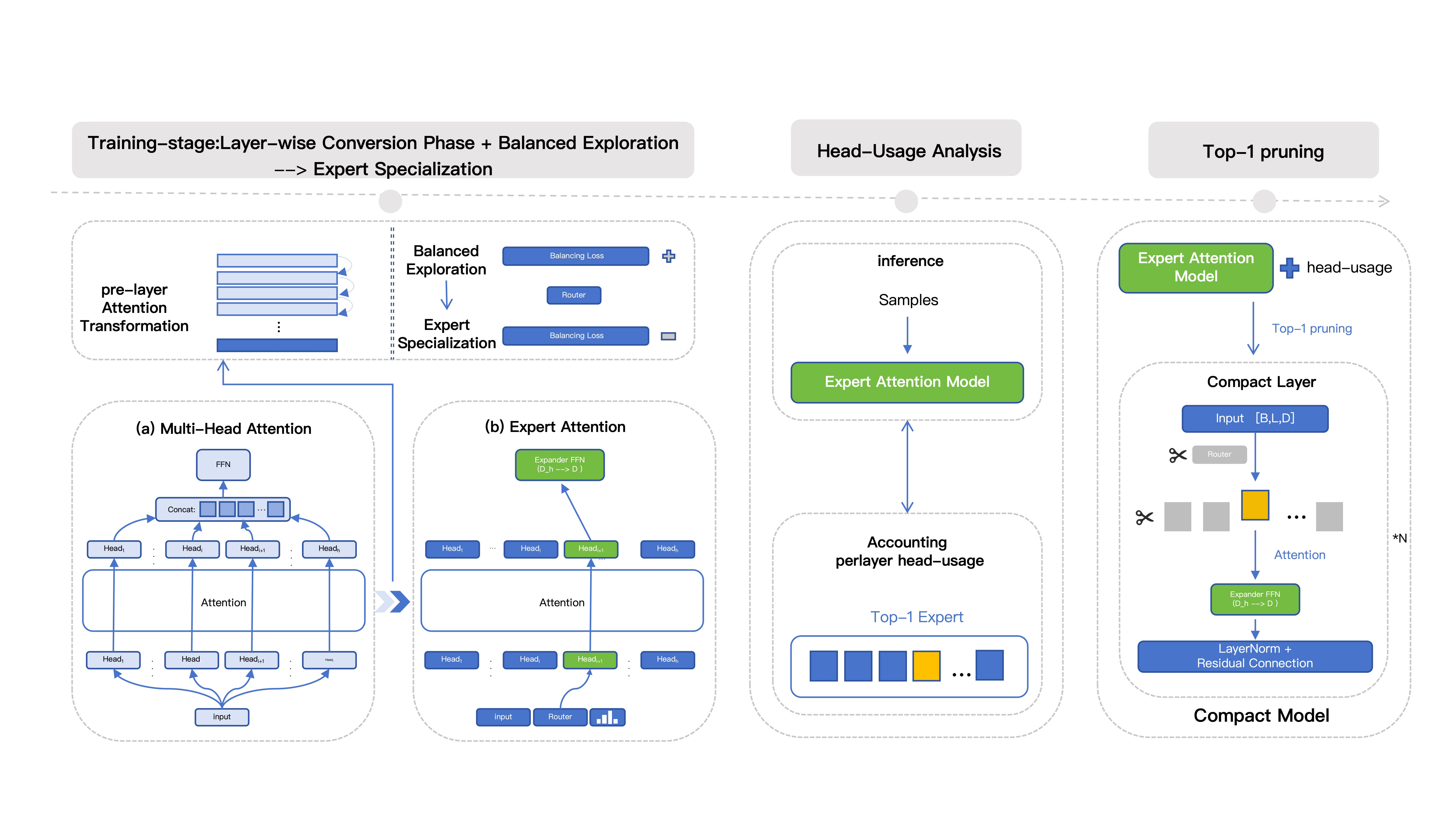}
    \caption{
    SHRP framework with progressive training, usage analysis, and Top-1 pruning.The framework consists of three key stages: (1) \emph{Training}, where standard multi-head attention is progressively transformed into \emph{Expert Attention} via layer-wise conversion; training proceeds in two phases—\emph{Balanced Exploration} (with load-balancing loss) and \emph{Expert Specialization} (with Top-1 routing); (2) \emph{Head-Usage Analysis}, where expert usage is measured per layer based on routing decisions during inference; (3) \emph{Top-1 Pruning}, where underused experts and routers are removed to produce a compact, deterministic encoder. Each expert head routes through a shared lightweight Expander FFN, enabling efficient structured compression across attention and FFN components.
    }
    \Description{Diagram of the SHRP framework showing progressive training, usage analysis, and Top-1 pruning.}
    \label{fig:shrp-pipeline}
\end{figure*}

\section{Related Work}

\paragraph{\textbf{Attention Head Pruning.}}  
Prior studies have shown that many attention heads in Transformer models are redundant. \citet{michel2019sixteen} demonstrated that large fractions of heads can be pruned without significant loss in accuracy. \citet{voita2019analyzing} introduced auxiliary losses to identify useful heads, while others proposed learnable masking or score-based heuristics~\cite{michel2019sixteen, DBLP}. While effective, these methods often require custom scoring functions or supervised signals, and typically focus only on the attention sublayer.

\textbf{\textit{Transformer Compression Techniques.}}  
Transformer compression has been widely explored to improve deployment efficiency. Coarse-grained methods like LayerDrop~\cite{fan2019reducing} and DynaBERT~\cite{hou2020dynabert} prune layers or adapt model depth/width dynamically. Fine-grained methods such as Movement Pruning~\cite{sanh2020movement} and quantization techniques~\cite{zafrir2019q8bert} operate at the parameter level but often introduce additional complexity and may require custom inference frameworks. Few of these methods offer structured, architecture-level pruning that preserves model modularity.

\textbf{\textit{Mixture-of-Experts (MoE) for Expansion.}}  
MoE architectures~\cite{shazeer2017outrageously, lepikhin2020gshard, fedus2022switch} scale model capacity via conditional computation, typically by routing tokens to subsets of FFN experts. These approaches enable large-scale training without proportional increases in compute, but are designed primarily for expansion, not compression. Moreover, most MoE methods target FFN blocks, leaving attention layers structurally untouched.

\textbf{\textit{MoE-Inspired Attention Head Routing.}}  
Recent work has explored applying MoE principles to attention heads. Notably, MoH~\cite{jin2024moh} reformulates multi-head attention as a dynamic mixture of heads, introducing a routing mechanism that selectively activates heads per token. MoH focuses on improving performance in vision and LLM contexts, and does not target structured pruning or inference efficiency. In contrast, our method reinterprets head modularity as a means for compression, not for architectural replacement.

\textbf{\textit{Our Contribution.}}  
In contrast to prior pruning and MoE approaches, our work targets \emph{structured compression} at the attention head level. SHRP leverages head modularity for pruning rather than expansion, and unifies routing and pruning under a Top-1 usage-driven mechanism. This design eliminates dynamic routing at inference while preserving efficiency, positioning SHRP as a deployment-oriented framework for compact and accurate Transformer encoders.

%% file: content/sec-3-Methodology.tex
\section{SHRP Framework} \label{sec:SHRP}
\paragraph{\textbf{Overview.}}

SHRP (\textit{Specialized Head Routing and Pruning}) is a structured pruning framework that reorganizes Transformer encoders around modularized attention heads. At a high level, SHRP replaces the standard multi-head attention block with \textbf{Expert Attention}, a design that treats each head as a specialized expert paired with a shared lightweight feed-forward network. This modularization enables two key operations: (1) \textit{usage-driven pruning}, where rarely activated experts can be removed cleanly, and (2) \textit{deterministic execution}, where only one expert is routed per input, eliminating dynamic routing overhead.

The framework proceeds in three stages: first, standard attention layers are converted into Expert Attention blocks; second, a two-stage training process encourages balanced expert utilization followed by specialization; finally, pruning is applied based on Top-1 usage statistics, yielding compact and deployment-ready encoder models. Figure~\ref{fig:shrp-pipeline} illustrates this overall workflow, including the transformation from standard attention to modular Expert Attention.

\subsection{From Standard Attention to Expert Attention}
SHRP builds on a simple but crucial change to the Transformer encoder: we replace the standard multi-head attention block with \textbf{Expert Attention}. In standard Transformers, all heads are computed jointly and their outputs are concatenated before passing through a large FFN. This coupling makes direct head pruning ineffective, as individual heads are not structurally independent.

\textbf{Expert Attention} addresses this by decoupling each attention head into an independent expert branch. Each branch computes its own query, key, and value projections and produces a head output that is passed through a shared lightweight Expander FFN. By isolating heads in this way, SHRP allows them to specialize during training and later be pruned cleanly without breaking the residual pathway.

This transformation process is illustrated in the \emph{Training-stage} section of Figure~\ref{fig:shrp-pipeline}. The figure contrasts the original multi-head attention structure (a), where all heads are jointly computed and passed through a large FFN, with the converted \emph{Expert Attention} design (b), in which each head operates as an independent expert routed through a shared lightweight Expander FFN. The conversion is introduced progressively across layers, accompanied by a two-phase training procedure—Balanced Exploration and Expert Specialization—which guides the heads toward diverse and stable specialization while preparing them for pruning.

\subsubsection{\textbf{Attention as Expert Module}}

In Expert Attention, we reinterpret each attention head as an independent expert module. Unlike standard multi-head attention, where heads are computed jointly and fused, our design detaches each head into a self-contained unit with its own query, key, and value projections:
\[
Q_i = XW^Q_i, \quad K_i = XW^K_i, \quad V_i = XW^V_i
\]
Each expert computes scaled dot-product attention individually:
\[
\text{Head}_i(X) = \text{softmax}\left( \frac{Q_i K_i^\top}{\sqrt{d_k}} \right) V_i
\]
These heads are not concatenated or merged. Instead, they act as standalone expert branches that will be selectively routed during inference.

By treating each head independently, we allow each to specialize without interference from others. This modularization also lays the foundation for subsequent pruning, as unused heads can be easily removed.

\subsubsection{\textbf{Expander FFN}}

In standard Transformer layers, the feed-forward network (FFN) is a two-layer MLP applied after the multi-head attention output, projecting from the hidden dimension $d$ to a larger intermediate dimension $d_{ff}$ and then back to $d$. This design introduces a significant number of parameters and dominates the overall model size.

In SHRP, we replace the original FFN with a simplified and shared \textbf{Expander FFN}, which directly projects from the head dimension $d_{\text{head}}$ to the model dimension $d$, followed by a lightweight non-linear transformation and normalization. Specifically, Expander FFN consists of a single linear projection, a GELU activation, and a LayerNorm:
\[
\text{ExpFFN}(x) = \text{LayerNorm}\big(\text{GELU}(W_{\text{exp}} x + b_{\text{exp}})\big),
\]
where $W_{\text{exp}} \in \mathbb{R}^{d \times d_{\text{head}}}$ and $b_{\text{exp}} \in \mathbb{R}^{d}$. 

This design serves three purposes:
\begin{enumerate}
    \item \textbf{Dimensional adaptation:} Each attention expert emits $(B, L, d_{\text{head}})$ with $d_{\text{head}} = d / h$. Expander FFN maps this directly back to $d$, ensuring compatibility with the residual path and subsequent normalization layers.
    \item \textbf{Functional substitution:} The combination of a linear projection, GELU, and LayerNorm replaces the original two-layer FFN, providing essential non-linear capacity and stable training dynamics.
    \item \textbf{Parameter efficiency:} Compared with the standard $d \to d_{ff} \to d$ FFN (with $2 d \cdot d_{ff}$ parameters), Expander FFN only requires $d \cdot d_{\text{head}}$ parameters for the linear projection (plus biases and normalization), leading to a drastic reduction in both parameter count and computation.
\end{enumerate}

As illustrated in Figure~\ref{fig:shrp-pipeline}, instead of aggregating all heads and applying a large two-layer FFN, SHRP decouples heads into experts and applies a shared lightweight Expander FFN, which efficiently restores dimensionality and injects non-linearity at a fraction of the cost.

\subsubsection{\textbf{Top-1 Gating}} \label{sec:top1gating}

We adopt a hard routing strategy that selects only a single expert per input ($k=1$). This design choice is driven purely by empirical observation: during pruning, we found that using multiple experts ($k > 1$) leads to significant accuracy degradation, while Top-1 routing consistently preserves performance.

The gating mechanism projects the [CLS] token representation $x_{\text{CLS}} \in \mathbb{R}^d$ to a score vector over $N$ experts:
\[
g = \text{Linear}(x_{\text{CLS}}) \in \mathbb{R}^N
\]
Then the top expert is selected as:
\[
i^* = \arg\max_j g_j
\]

Only the selected expert $i^*$ is executed; no softmax or weighting is applied. This simple routing not only reduces computation but also produces clear usage patterns, making expert pruning straightforward.

We validate the impact of this choice in Section~\ref{sec:top1-ablation}, where we compare Top-$k$ routing variants under different pruning levels.

\subsubsection{\textbf{Integration and Output}}

As illustrated in first stage of Figure~\ref{fig:shrp-pipeline}, once the top-1 expert $i^*$ is selected, its attention head processes the input and passes the result through the shared Expander FFN:
\[
E = \text{ExpFFN}(\text{Head}_{i^*}(X))
\]

We then apply residual connection and layer normalization, using the original input $X$:
\[
Y = \text{LayerNorm}(E + X)
\]

This design ensures that only a single expert is executed per layer and per input. Because experts are fully decoupled, those that are rarely or never selected can be pruned entirely—enabling efficient post-training compression without any structural reconfiguration.

\subsection{Progressive Training for Expert Specialization}

Training an Expert Attention model poses unique challenges due to the sparse and modular nature of its computation. In particular, directly applying MoE transformation to all layers at once often leads to instability or training collapse, especially when fine-tuning on downstream tasks. To address this, we adopt a progressive training strategy with two critical components: \textbf{layer-wise MoE conversion} and a \textbf{two-stage optimization schedule}.

\paragraph{Layer-wise MoE Conversion.}
Instead of transforming all encoder layers into Expert Attention modules simultaneously, we convert one layer at a time in a staged manner. At each training epoch, one additional Transformer layer is replaced by its MoE counterpart. Once all target layers have been converted, we continue training the full model for a small number of extra epochs (typically 2) to stabilize learning and allow for full adaptation.

For example, when training a 12-layer encoder with 6 MoE-converted layers over 8 epochs, we proceed as follows:
\begin{itemize}
    \item Epochs 1–6: progressively convert 1 new layer per epoch.
    \item Epochs 7–8: full MoE model trained as-is for final optimization.
\end{itemize}
This approach prevents sudden optimization shocks, allowing each newly introduced expert layer to adapt gradually to the model's dynamics.

\paragraph{Two-Stage Learning: Load Balancing $\rightarrow$ Specialization.}
We further split the training process into two functional stages to align with the final goal of pruning underutilized experts:

\begin{itemize}
    \item \textbf{Stage 1 - Balanced Exploration.}  
    During the layer-wise conversion phase (before all target layers are MoE-enabled), we apply a load balancing loss to encourage even expert utilization. This prevents early expert collapse and ensures that each expert receives sufficient gradient signals.

    Unlike simple entropy-based approaches, we compute the load balancing loss via bidirectional KL divergence between the observed expert usage distribution and an ideal uniform prior. For each MoE layer, let $p \in \mathbb{R}^N$ denote the mean routing probability vector over a batch for $N$ experts (computed as $\text{softmax}(\text{router\_logits})$.mean(dim=0)), and let $u = \frac{1}{N} \cdot \mathbf{1}$ be the uniform target distribution. To ensure numerical stability, we add a small $\epsilon = 10^{-7}$ to both distributions.

    The load balancing loss is defined as:
    \begin{align*}
    \mathcal{L} &= \frac{1}{2} \cdot \text{KL}\left( \log(p + \epsilon) \,\|\, u + \epsilon \right) \\
                &\quad + \frac{1}{2} \cdot \text{KL}\left( \log(u + \epsilon) \,\|\, p + \epsilon \right)
    \end{align*}

    where $\text{KL}(\cdot \,\|\, \cdot)$ denotes the Kullback-Leibler divergence. For models with multiple MoE layers, we average the load balancing loss across all layers.

    The total training loss in this stage becomes:
    \[
    \mathcal{L}_{\text{total}} = \mathcal{L}_{\text{task}} + \lambda \cdot \mathcal{L}_{\text{balance}}
    \]

    where $\lambda$ is a fixed coefficient, set to 0.1 in our implementation.

    Once all target layers are converted to MoE layers (i.e., when $\text{current\_modified\_layers} \geq \text{target\_modified\_layers}$), we disable $\mathcal{L}_{\text{balance}}$ to allow the gating mechanism to naturally specialize experts in Stage 2.

    \item \textbf{Stage 2 - Expert Specialization.}  
    After all MoE layers are in place, we disable the load balancing objective and allow the gating mechanism to focus solely on task performance. In this phase, experts naturally specialize, and frequently selected experts are reinforced, while unselected ones begin to fade. This directly supports the goal of pruning, where underused experts can later be removed without harming performance. Importantly, the Top-1 gating strategy is preserved throughout to ensure sparsity and modularity.
\end{itemize}

This two-phase training process bridges the gap between robust early learning and efficient late-stage specialization. It ensures both high-quality optimization and structural sparsity, laying the groundwork for expert pruning in subsequent steps.

\subsection{Top-1 Pruning} \label{pruning}
Building on the Top-1 gating mechanism described in Section~\ref{sec:top1gating}, 
we perform pruning based directly on the observed usage frequencies. 
Since each input is routed to exactly one expert during training, the gating decisions 
produce clear selection statistics, enabling a simple usage-driven pruning strategy. 
Experts that are rarely or never selected are pruned entirely, yielding a compact 
and efficient encoder.

\paragraph{Usage-based Expert Selection.}
Each MoE layer routes inputs to one of several expert heads based on a Top-1 gating decision. After training, we analyze validation-time expert usage across all layers by recording how frequently each expert is selected. For each layer, we rank experts by their activation frequency and identify those that contribute meaningfully to inference.

\paragraph{Pruning Policy.}
For every MoE layer, we retain only the top-$m$ most frequently used experts and discard the rest. The choice of $m$ can be tuned to control the trade-off between model compression and performance retention. Since our architecture uses hard routing (Top-1) and each expert is structurally independent, the unused branches can be safely removed without any additional fine-tuning.

\paragraph{Architecture Simplification.}
Following pruning, we statically replace each dynamic MoE layer with a simplified deterministic attention layer that only includes the selected experts. This change eliminates the runtime cost of gating and makes inference more predictable and hardware-efficient.

\paragraph{Effectiveness.}
This approach yields substantial reductions in parameter count and inference latency. Unlike unstructured pruning, our method operates at the architectural level—making the resulting model easy to deploy and analyze. Experimental results show that pruning has minimal impact on downstream task performance while significantly improving throughput and efficiency.

%% file: content/sec-4-Experiment.tex
\section{Experiments}

We evaluate \textbf{SHRP} with the goal of answering three key research questions:

\begin{itemize}
    \item \textbf{RQ1: Task performance under compression.} 
    How does our method perform across GLUE tasks when compared to existing model compression and pruning techniques, in terms of task accuracy, parameter reduction, and inference speedup?
    
    \item \textbf{RQ2: Comparison with prior approaches.} 
    How does SHRP compare with existing attention head pruning methods in terms of accuracy, compression, and inference efficiency, especially in the context of deployment-ready systems for Web- and mobile-scale inference?

    \item \textbf{RQ3: Why Top-1 routing and pruning?}
    We empirically explore Top-$K$ gating configurations to determine which $K$ yields the best trade-off between expert specialization and post-pruning performance, and to explain why the Top-1 gating + pruning strategy is most appropriate for structured compression.

\end{itemize}

\begin{table*}[ht]
\caption{Performance of different pruning degrees on seven GLUE tasks. 
All evaluation metrics are reported as accuracy. 
The first column indicates the pruning ratio (\emph{pruned layers / total layers}). 
The last four columns summarize the averaged changes across tasks: parameter reduction ratio (excluding embedding layers), performance retention ratio, throughput improvement ratio compared with the baseline, and the percentage of FLOPs remaining relative to the baseline model.}

\centering
\small
\renewcommand{\arraystretch}{1.30}
\setlength{\tabcolsep}{3.6pt}
\setlength\dashlinedash{1.5pt}
\setlength\dashlinegap{1.5pt}
\newcolumntype{Y}{>{\centering\arraybackslash}p{19mm}}
\newcommand{\trow}{\rule{0pt}{2.6ex}}

\begin{tabular}{c:ccccccc:YYYY}
\hline
\multirow{2}{*}{\textbf{Pruning}}
& \multicolumn{7}{c:}{\textbf{GLUE Task Scores (Acc)}} 
& \multirow{2}{*}{\makecell{\textbf{Param}\\\textbf{Reduction} (\%)}} 
& \multirow{2}{*}{\makecell{\textbf{Performance}\\\textbf{Retention} (\%)}} 
& \multirow{2}{*}{\makecell{\textbf{Throughput}\\$\uparrow$ (\%)}} 
& \multirow{2}{*}{\makecell{\textbf{FLOPs}\\\textbf{Remaining} (\%)}} \\
\cdashline{2-8}[1pt/1pt]
& \textbf{MNLI} & \textbf{QQP} & \textbf{QNLI} & \textbf{SST-2} & \textbf{MRPC} & \textbf{RTE} & \textbf{CoLA} & & & & \\
\hline
\rowcolor{gray!15}\arrayrulecolor{black}
Baseline \trow
& 0.8499 & 0.9098 & 0.9081 & 0.9209 & 0.8480 & 0.5776 & 0.8226
& 0 & 100 & 0 & 100 \\
\hdashline
2/12  \trow
& 0.8501 & 0.9086 & 0.8960 & 0.9255 & 0.8162 & 0.6173 & 0.8111
& 16.1 & 100.1 & 18.7 & 83.91 \\
4/12  \trow
& 0.8409 & 0.9071 & 0.8882 & 0.9151 & 0.8039 & 0.5993 & 0.7910
& 32.2 & 98.6 & 43.0 & 67.82 \\
6/12  \trow
& 0.8241 & 0.9050 & 0.8755 & 0.9002 & 0.6838 & 0.5560 & 0.7248
& 48.2 & 93.1 & 81.1 & 51.74 \\
8/12  \trow
& 0.7949 & 0.8955 & 0.8444 & 0.8739 & 0.7034 & 0.4801 & 0.6836
& 64.3 & 89.2 & 146.2 & 35.65 \\
10/12 \trow
& 0.7468 & 0.8755 & 0.8175 & 0.8406 & 0.6838 & 0.5523 & 0.6884
& 80.4 & 89.6 & 282.1 & 19.56 \\
11/12 \trow
& 0.6853 & 0.8572 & 0.6096 & 0.8337 & 0.6838 & 0.5307 & 0.6740
& 88.5 & 84.4 & 424.1 & 11.52 \\
\hline
\end{tabular}

\label{tab:glue_fixed}
\end{table*}

\begin{figure*}
    \centering
    \includegraphics[width=\textwidth]{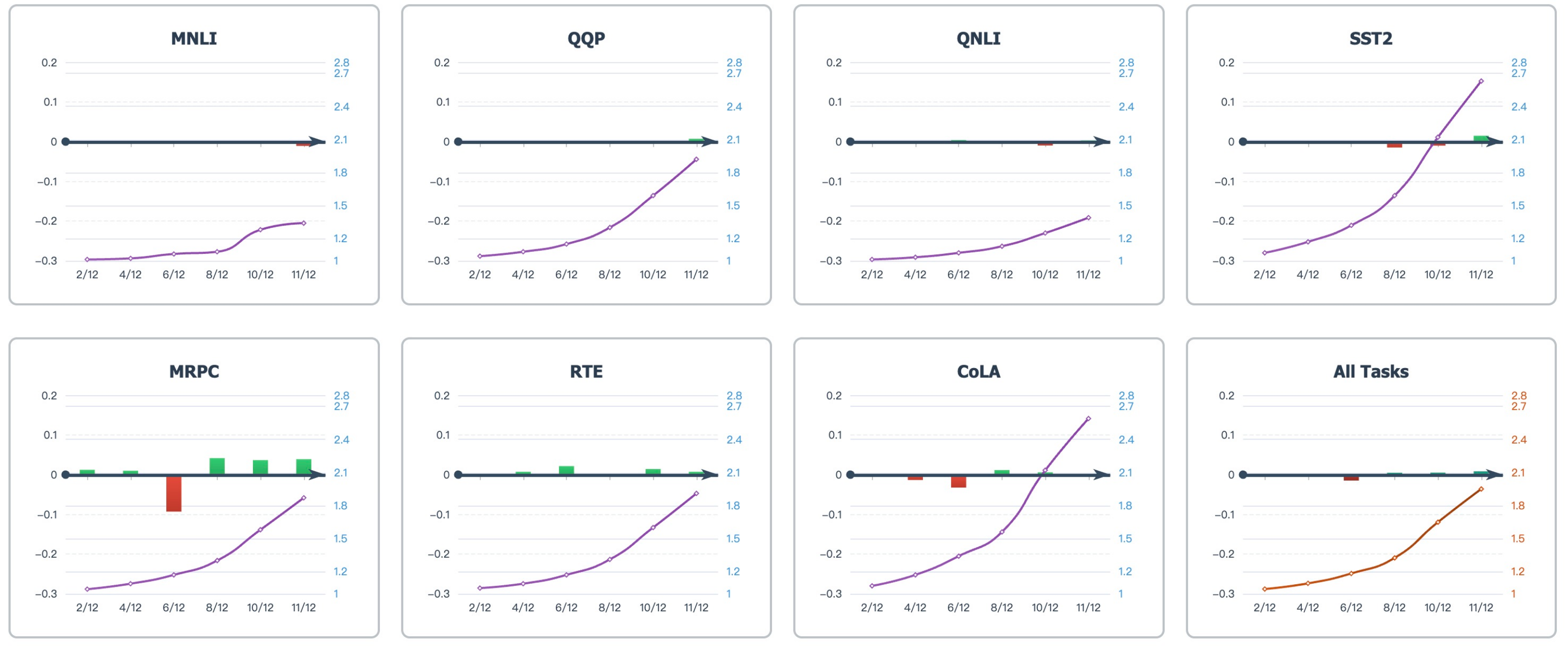}
    \caption{
    Task-wise visualization of pruning impact across the GLUE benchmark. 
    Each subplot corresponds to a task, showing the effect of SHRP pruning at different layer ratios (2/12 to 11/12).
    The bar chart (left Y-axis) represents the change in task accuracy compared to the baseline: upward green bars indicate improved performance post-pruning, while downward red bars reflect performance drops. 
    The purple line (right Y-axis) shows the corresponding throughput gain (x-fold), measured without the gating router overhead. 
    The final panel aggregates results across all tasks, highlighting the average performance change and overall efficiency gains.
    }
    \label{fig:placeholder}
\end{figure*}

\subsection{Experimental Settings}

\paragraph{\textbf{Datasets.}} 
We evaluate our framework on the \textbf{GLUE} benchmark~\cite{warstadt2018neural,wang2019glue}, a standard suite for natural language understanding. 
We consider seven representative tasks: MNLI, QQP, QNLI, SST-2, MRPC, RTE, and CoLA. 
These tasks collectively cover natural language inference, paraphrase detection, sentiment classification, and linguistic acceptability, providing a diverse and challenging evaluation setup. 
All experiments use \texttt{BERT-base-uncased} (12 layers, 12 heads per layer) as the default backbone unless otherwise specified.

\paragraph{\textbf{Baselines.}} 
We benchmark our method against several representative attention head pruning techniques widely adopted in prior literature~\cite{DBLP}. 
All baselines are evaluated with the same number of retained heads to ensure a fair comparison of pruning strategies. 

The baseline methods include:
\begin{itemize}
    \item \textbf{Michel et al.}~\cite{base_1}: Prunes attention heads based on gradient-derived importance scores, requiring backpropagation-based sensitivity analysis to identify redundant heads. 
    \item \textbf{Voita et al.}~\cite{base_3}: Applies auxiliary loss functions during training to guide binary head selection, which increases training complexity and introduces additional hyperparameters. 
    \item \textbf{Pipelined DSP} and \textbf{Joint DSP}~\cite{base_2,DBLP}: Employ iterative and end-to-end optimization strategies to dynamically remove redundant heads, but rely on heavy optimization procedures that are difficult to reproduce and scale. 
    \item \textbf{STE}~\cite{base_4}: Uses a straight-through estimator for differentiable head masking, allowing gradient-based pruning but introducing approximation error and instability during training. 
\end{itemize}

While effective at reducing attention redundancy, prior methods have key limitations. They often rely on complex importance scores or masking schemes, making them less transparent and harder to generalize. More critically, they focus only on pruning attention heads while leaving FFNs intact, limiting overall compression and speedup. SHRP addresses this by modularizing the entire Transformer block: each attention head is paired with a lightweight FFN, enabling consistent pruning across both sublayers. This leads to simpler training, higher compression, and models better suited for real-world inference.

\textbf{Implementation details.} All experiments are conducted on a single NVIDIA A800 GPU (48GB memory). 
Each GLUE task exclusively occupies one GPU to avoid interference when measuring throughput. 
We adopt the following fine-tuning protocol: 
\begin{itemize}
    \item \textbf{Optimizer:} AdamW with learning rate $2{\times}10^{-5}$, weight decay $0.01$, and gradient clipping at max-norm 1.0. 
    \item \textbf{Learning rate schedule:} cosine annealing with a 10\% warmup ratio, i.e., warmup steps equal to 10\% of total training steps. 
    \item \textbf{Batch size:} 64, following standard GLUE fine-tuning practice. 
\end{itemize}

\textbf{Pruning protocol.}
We progressively replace $Z$ out of 12 encoder layers with \textbf{Expert Attention} modules, using the two-stage training schedule described in Section~3.1 (balanced exploration $\rightarrow$ specialization). 
During training, a lightweight Top-1 router records expert usage at each layer. 
After convergence, we apply \textbf{Top-1 pruning} by removing underutilized experts entirely and discarding the router. 
This produces a compact, deterministic encoder that is free of dynamic routing overhead at inference time.

\textbf{Evaluation metrics.}
We evaluate models on three dimensions: 
(1) \textit{Task accuracy} ; 
(2) \textit{Compression ratio}, measured as the percentage of parameters pruned (excluding embeddings); 
and (3) \textit{Inference efficiency}, measured by throughput (samples/second, batch size 64) and latency (milliseconds/sample). 
These metrics directly reflect the requirements of \textbf{deployment-ready systems} for Web and mobile inference, where both efficiency and accuracy are critical.

\begin{table*}[ht]
\caption{Accuracy comparison of SHRP with existing attention head pruning methods on MNLI. We report accuracy under different numbers of unpruned heads, along with corresponding model size reduction (baseline methods. vs ours.).}

\centering
\resizebox{\textwidth}{!}{
\begin{tabular}{m{1.7cm}<{\centering} m{1.7cm}<{\centering} m{1.7cm}<{\centering} m{1.7cm}<{\centering} m{1.7cm}<{\centering} m{1.7cm}<{\centering} m{2cm}<{\centering} m{3cm}<{\centering}}
\hline
\textbf{Unpruned Heads} & \textbf{Michel et al.} & \textbf{Pipelined DSP} & \textbf{Voita et al.} & \textbf{STE} & \textbf{Joint DSP} & \textbf{SHRP (Ours)} & \textbf{Model Size Reduction} \\
\hline
120 & 0.8046 & 0.8441 & 0.8418 & 0.8459 & 0.8497 & \textbf{0.8501} & 4.0 / \textbf{16.6} \\
96 & 0.8424 & 0.8327 & 0.8295 & 0.8393 & 0.8441 & 0.8409 & 8.0 / \textbf{32.6} \\
72 & 0.8247 & 0.8295 & 0.8324 & 0.8281 & 0.8348 & 0.8241 & 13.0 / \textbf{48.6} \\
48 & 0.7926 & 0.7910 & 0.7608 & 0.8231 & 0.8322 & 0.7949 & 17.0 / \textbf{64.5} \\
24 & 0.7082 & 0.7629 & 0.3168 & 0.8220 & 0.8251 & 0.7468 & 19.0 / \textbf{80.5} \\
12 & 0.4059 & 0.5629 & 0.7691 & 0.7379 & 0.7974 & 0.6853 & 24.0 / \textbf{88.5} \\
\hline
\end{tabular}
}

\label{table:pruning_comparison}
\end{table*}

\begin{table*}[!htbp]
\caption{Layer-wise accuracy differences (\textit{post-prune minus pre-prune}) on the QQP task under different Top-$k$ routing and pruning settings. Results are reported for pruned layers only. The final column shows the average accuracy difference across all pruned layers. Positive values indicate that pruning improved performance, while negative values denote accuracy degradation.}

\centering
\setlength{\tabcolsep}{6pt}
\renewcommand{\arraystretch}{1.1}
\begin{tabular}{c|cccccc|c}
\hline
\textbf{Top-$k$} & \textbf{Layer 2} & \textbf{Layer 4} & \textbf{Layer 6} & \textbf{Layer 8} & \textbf{Layer 10} & \textbf{Layer 11} & \textbf{Avg Diff} \\
\hline
1  & +0.0011 & +0.0011 & +0.0023 & -0.0046 & +0.0149 & -0.0023 & +0.0021 \textcolor{green}{\textbf{(↑)}} \\
3  & -0.0080 & -0.2534 & -0.3498 & -0.3394 & -0.2706 & -0.2729 & -0.2490 \textcolor{red}{(↓)} \\
5  & -0.0264 & -0.2626 & -0.3567 & -0.3567 & -0.3291 & -0.3337 & -0.2775 \textcolor{red}{(↓)} \\
7  & -0.0390 & -0.2913 & -0.3624 & -0.3486 & -0.3165 & -0.3326 & -0.2817 \textcolor{red}{(↓)} \\
9  & -0.0195 & -0.2890 & -0.3463 & -0.3589 & -0.3314 & -0.3165 & -0.2769 \textcolor{red}{(↓)} \\
11 & -0.0677 & -0.3291 & -0.3784 & -0.3624 & -0.3578 & -0.3280 & -0.3039 \textcolor{red}{(↓)} \\
\hline
\end{tabular}

\label{tab:topk_diff_avg_noempty}
\end{table*}

\subsection{Experimental Results}

\subsubsection*{4.2.1 Performance under Different Pruning Ratios}

Table~\ref{tab:glue_fixed} shows the impact of varying pruning ratios across seven GLUE tasks. SHRP demonstrates strong robustness under compression, consistently preserving task accuracy while significantly reducing parameters, FLOPs, and inference latency. Key observations:

\begin{itemize}[leftmargin=1.5em, itemsep=2pt]
    \item \textbf{Moderate pruning offers an excellent trade-off.} With 6 out of 12 layers pruned, SHRP achieves a 48.2\% parameter reduction, 51.7\% FLOPs reduction, and an 81.1\% throughput gain, while retaining 93.1\% of the baseline accuracy.
    
    \item \textbf{Even aggressive pruning remains competitive.} At 10/12 pruning, the model maintains nearly 90\% task performance with 80.4\% fewer parameters, 19.6\% of original FLOPs, and a 2.8$\times$ speedup. At 11/12 pruning, SHRP reduces parameters by 88.5\%, FLOPs by 11.5\%, and still retains 84.4\% accuracy with a 4.2$\times$ throughput gain.
    
    \item \textbf{Graceful degradation.} Unlike many pruning approaches that collapse under extreme compression, SHRP exhibits a predictable trade-off between accuracy and efficiency across all pruning levels.
\end{itemize}

These results confirm that SHRP's expert-based pruning delivers highly compact, performant models that are well suited for real-world Web and mobile deployment.

\subsubsection{Task-wise Impact and Throughput Gains}

Figure~\ref{fig:placeholder} provides a task-level breakdown of pruning effects.  
The results highlight two major findings:

\begin{itemize}
    \item \textbf{Accuracy robustness.}  
    Across the majority of GLUE tasks, pruning introduces only \textbf{negligible accuracy loss} (typically within $0.03$).  
    In some cases (e.g., MNLI and SST-2), pruning even improves generalization, suggesting that redundant heads may introduce noise rather than useful capacity.  
    This demonstrates that our \textbf{two-stage training schedule} effectively encourages heads to specialize, enabling safe pruning.
    
   \item \textbf{Substantial throughput improvements.}  
    As the pruning ratio increases, more routers are eliminated, leading to progressively larger speed gains.  
    Even at a light pruning ratio of 2/12, throughput improves by at least \textbf{1.1$\times$} compared to the baseline.  
    The effect becomes more pronounced under aggressive pruning: for example, on SST-2 with 11/12 layers pruned, 
    SHRP achieves nearly a \textbf{2.7$\times$ speedup}.  
    By \textbf{removing the router entirely at inference}, the model avoids dynamic routing overhead 
    and converges to a \textbf{deterministic, lightweight encoder}, well-suited for deployment.

\end{itemize}

Taken together, these findings underscore that \textbf{SHRP not only compresses models effectively but also delivers deployment-friendly inference speedups}, validating the necessity of structured compression in modern Transformer systems.

\subsubsection{Comparison with Traditional Pruning Methods}
Table~\ref{table:pruning_comparison} compares \textbf{SHRP} with several representative attention head pruning methods on MNLI under varying numbers of retained heads. 
All approaches are evaluated with the same pruning budget, ensuring a fair comparison of pruning strategies. 
The results highlight three key findings:
\begin{itemize}
    \item \textbf{Higher accuracy at equivalent pruning.}  
    Across most pruning levels, SHRP achieves accuracy that is on par with or better than existing methods. For example, with 120 or 96 retained heads, our approach reaches \textbf{0.8501 and 0.8409 accuracy}, matching or slightly exceeding the best-performing baselines while providing stronger compression.
    \item \textbf{Much stronger compression at aggressive pruning.}  
    The advantage of SHRP becomes most evident under extreme compression.  
    With only 12 retained heads, our method achieves an \textbf{88.5\% parameter reduction}, compared to just \textbf{24\% in Joint DSP}, while still maintaining competitive accuracy (\textbf{0.6853 vs. 0.7974}).  
    This demonstrates that SHRP can prune much more aggressively while preserving usable performance.
    \item  \textbf{Better efficiency–accuracy trade-off.}  
    Unlike traditional methods that prune only attention heads while leaving FFNs untouched, SHRP jointly prunes both attention and FFN sublayers via its modular design.  
    This not only reduces parameter counts more substantially but also translates to \textbf{real-world throughput gains}, as validated by our hardware-level measurements (see Appendix Table~\ref{table:pruning_performance}).
\end{itemize}
Collectively , these results demonstrate that \textbf{SHRP consistently outperforms traditional head pruning techniques in compression ratio and runtime efficiency}, without sacrificing accuracy at moderate pruning levels.  In aggressive settings, our approach is particularly advantageous, delivering deployment-ready models that are both compact and efficient.

\subsection{Top-1 Ablation Study} \label{sec:top1-ablation}

To better understand the effect of routing configurations on pruning effectiveness, 
we conduct an ablation study on QQP by comparing \textbf{Top-1 routing} with larger Top-$k$ settings 
($k \in \{3,5,7,9,11\}$). After applying pruning across 2/12–11/12 layers, 
we measure the post-prune accuracy differences relative to the unpruned baseline. 
The detailed layer-wise results are shown in Table~\ref{tab:topk_diff_avg_noempty}.Three key observations emerge from the results:

\begin{itemize}
    \item \textbf{Top-1 routing is the only stable configuration.}  
    Across all pruned layers, Top-1 shows accuracy changes fluctuating narrowly around zero 
    (e.g., +0.0011 at Layer 2, +0.0149 at Layer 10), resulting in a small but positive average gain 
    of \textbf{+0.0021}. This indicates that pruning under Top-1 can even slightly improve performance 
    by removing redundant experts.  
    \item \textbf{Larger $k$ leads to consistent degradation.}  
    For Top-3 to Top-11, every layer shows negative accuracy differences, with losses growing 
    steadily as $k$ increases (e.g., Layer 6: $-0.3498$ at Top-3 vs. $-0.3784$ at Top-11).  
    The averaged drop reaches up to \textbf{-0.3039}, confirming that larger $k$ consistently harms pruning performance.
    \item \textbf{Interpretability of usage statistics.}  
    Top-1 enforces a winner-takes-all mechanism, producing \textbf{sparse and reliable usage counts} 
    that make it straightforward to identify underutilized experts for pruning.  
    By contrast, higher $k$ encourages expert overlap and parameter interference, 
    leading to noisy selection patterns and less effective pruning outcomes.
\end{itemize}

\textit{Cross-task Robustness.} 
Although the analysis above is shown for QQP, we observe the same trend across other GLUE tasks.  
Top-1 routing yields \textbf{stable accuracy retention and consistent throughput improvements}, 
while larger $k$ values suffer from significant and systematic degradation.  
This stability highlights Top-1 as the most practical choice for deployment-ready pruning.

\textbf{In summary}, Top-1 routing offers the best balance of accuracy and compression. It minimizes performance loss, yields interpretable expert usage, and enables compact, deterministic models without routing overhead. These results validate Top-1 as the most effective and explainable pruning strategy in SHRP. Full ablation results are in Appendix Table~5.

%% file: content/appendix.tex

\clearpage
\appendix

\section{Why Encoder-Only Models}
\label{appendix:why-encoders}

While the SHRP framework can, in principle, be extended to decoder-based or encoder–decoder architectures, 
we focus on encoder-only models in this work for three practical and methodological reasons:

\begin{enumerate}
    \item \textbf{Parallel inference and bidirectional context.} 
    Encoder models perform full-sequence processing in parallel and capture bidirectional dependencies, 
    which are crucial for tasks such as classification, retrieval, and reranking. 
    This property makes them ideal candidates for structured compression, 
    where latency and throughput improvements directly translate to production efficiency.

    \item \textbf{Stability under structured pruning.}
    Decoder-only models, in contrast, rely on autoregressive token-by-token generation, 
    where pruning or routing decisions can interact with temporal dependencies 
    and amplify instability across decoding steps. 
    Structured pruning in this setting often introduces unpredictable latency and quality degradation, 
    whereas encoders offer deterministic computation paths and stable optimization behavior.

    \item \textbf{Dominance in real-world deployment.} 
    In large-scale applications such as web ranking, content moderation, and recommendation, 
    encoder-based architectures are widely adopted due to their deterministic runtime, 
    memory predictability, and compatibility with high-throughput serving infrastructure.
    Improvements in encoder efficiency therefore have immediate impact on industrial-scale deployments.
\end{enumerate}

In summary, encoders strike a balance between representational expressiveness and operational determinism, 
making them the most practical and impactful testbed for exploring structured compression via SHRP.


\begin{table*}[!htbp]
\caption{Performance comparison of pruning methods in terms of throughput (batch size 64), accuracy, and model size reduction. The first two rows are baselines (no pruning; extreme head pruning), followed by MoE-based layer pruning with progressively fewer unpruned heads per layer.}

\centering
\setlength{\tabcolsep}{8pt}
\renewcommand{\arraystretch}{1.12}
\resizebox{\textwidth}{!}{
\begin{tabular}{c c m{3cm}<{\centering} c m{4cm}<{\centering}}
\hline
\textbf{Model (Unpruned Heads)} & \textbf{Throughput (inf/s)} & \textbf{Model Size Reduction (\%)} & \textbf{Acc} & \textbf{Accuracy Retained (\%)} \\
\hline
bert-base-uncased (144) & 1500 & 0 & 84.90 & / \\
\hdashline
Joint DSP (12) & $\approx$ 1995 (↑33\%) & -26\% & 61.79 & 72.78\% \\
\hdashline
layer 2/12 (120) & 1780.50 (↑18.7\%) & 16.60\% & \textbf{85.01} &  \textbf{100.12\%} \\
layer 4/12 (96)  & 2145.00 (↑43.0\%) & -32.60\% & 84.09 & 99.05\% \\
layer 6/12 (72)  & 2716.50 (↑81.1\%) & -48.60\% & 82.41 & 97.14\% \\
layer 8/12 (48)  & 3219.30 (↑146.2\%) & -64.50\% & 79.49 & 93.63\% \\
layer 10/12 (24) & 5731.50 (↑282.1\%) & -80.50\% & 74.68 & 87.97\% \\
layer 11/12 (12) & \textbf{7861.50 (↑424.1\%)} & \textbf{-88.50\%} & 68.53 & 81.11\% \\
\hline
\end{tabular}
}

\label{table:pruning_performance}
\end{table*}

\FloatBarrier

\begin{table*}[!htbp]
\caption{Detailed accuracy changes under different Top-$k$ routing settings and pruning depths (2--11 out of 12 layers) on QQP. Each entry reports accuracy before and after pruning, along with the difference.}

\centering
\setlength{\tabcolsep}{8pt}
\renewcommand{\arraystretch}{1.15}
\begin{tabular}{c|c|c|c|c}
\hline
\textbf{Top-$k$} & \textbf{Pruned Layers} & \textbf{Pre-Prune Acc} & \textbf{Post-Prune Acc} & \textbf{Diff} \\
\hline
\multirow{6}{*}{1}
& 2  & 0.9174 & 0.9186 & +0.0011 \\
& 4  & 0.9094 & 0.9106 & +0.0011 \\
& 6  & 0.9071 & 0.9094 & +0.0023 \\
& 8  & 0.8922 & 0.8876 & $-$0.0046 \\
& 10 & 0.8429 & 0.8578 & +0.0149 \\
& 11 & 0.8188 & 0.8165 & $-$0.0023 \\
\hline
\multirow{6}{*}{3}
& 2  & 0.9209 & 0.9128 & $-$0.0080 \\
& 4  & 0.9163 & 0.6628 & $-$0.2534 \\
& 6  & 0.8968 & 0.5470 & $-$0.3498 \\
& 8  & 0.8853 & 0.5459 & $-$0.3394 \\
& 10 & 0.8463 & 0.5757 & $-$0.2706 \\
& 11 & 0.8337 & 0.5608 & $-$0.2729 \\
\hline
\multirow{6}{*}{5}
& 2  & 0.9266 & 0.9002 & $-$0.0264 \\
& 4  & 0.9128 & 0.6502 & $-$0.2626 \\
& 6  & 0.9002 & 0.5436 & $-$0.3567 \\
& 8  & 0.8865 & 0.5298 & $-$0.3567 \\
& 10 & 0.8544 & 0.5252 & $-$0.3291 \\
& 11 & 0.8429 & 0.5092 & $-$0.3337 \\
\hline
\multirow{6}{*}{7}
& 2  & 0.9209 & 0.8819 & $-$0.0390 \\
& 4  & 0.9094 & 0.6181 & $-$0.2913 \\
& 6  & 0.9037 & 0.5413 & $-$0.3624 \\
& 8  & 0.8784 & 0.5298 & $-$0.3486 \\
& 10 & 0.8578 & 0.5413 & $-$0.3165 \\
& 11 & 0.8394 & 0.5069 & $-$0.3326 \\
\hline
\multirow{6}{*}{9}
& 2  & 0.9255 & 0.9060 & $-$0.0195 \\
& 4  & 0.9128 & 0.6239 & $-$0.2890 \\
& 6  & 0.9025 & 0.5562 & $-$0.3463 \\
& 8  & 0.8979 & 0.5390 & $-$0.3589 \\
& 10 & 0.8521 & 0.5206 & $-$0.3314 \\
& 11 & 0.8475 & 0.5310 & $-$0.3165 \\
\hline
\multirow{6}{*}{11}
& 2  & 0.9163 & 0.8486 & $-$0.0677 \\
& 4  & 0.9140 & 0.5849 & $-$0.3291 \\
& 6  & 0.9014 & 0.5229 & $-$0.3784 \\
& 8  & 0.8888 & 0.5264 & $-$0.3624 \\
& 10 & 0.8578 & 0.5000 & $-$0.3578 \\
& 11 & 0.8326 & 0.5046 & $-$0.3280 \\
\hline
\end{tabular}

\label{tab:topk_detailed}
\end{table*}

\FloatBarrier